# Machine Learning-based EEG Applications and Markets

Weiqing Gu, Bohan Yang, Ryan Chang

gu@dasion.ai

by93@cornell.edu

rchang05@usc.edu

## Abstract：

This paper addresses both the various EEG applications and the current EEG market ecosystem propelled by machine learning. Increasingly available open medical and health datasets using EEG encourage data-driven research with a promise of improving neurology for patient care through knowledge discovery and machine learning data science algorithm development. This effort leads to various kinds of EEG developments and currently forms a new EEG market. This paper attempts to do a comprehensive survey on the EEG market and covers the six significant applications of EEG, including diagnosis/screening, drug development, neuromarketing, daily health, metaverse, and age/disability assistance. The highlight of this survey is on the compare and contrast between the research field and the business market. Our survey points out the current limitations of EEG and indicates the future direction of research and business opportunity for every EEG application listed above. Based on our survey, more research on machine learning-based EEG applications will lead to a more robust EEG-related market. More companies will use the research technology and apply it to real-life settings. As the EEG-related market grows, the EEG-related devices will collect more EEG data, and there will be more EEG data available for researchers to use in their study, coming back as a virtuous cycle. Our market analysis indicates that research related to the use of EEG data and machine learning in the six applications listed above points toward a clear trend in the growth and development of the EEG ecosystem and machine learning world.

# I. Introduction

This paper gives an essential background about EEG in *section I*, including EEG bands, early history, EEG acquisition process, quantitative EEG, advantages and disadvantages, and EEG applications. Then, we summarized the most recent papers since 2017 about the six significant applications of EEG in *section II*. The primary application includes diagnosis/screening, drug development, neuromarketing, daily health, disability assistance, and the metaverse. In *section III*, we collected various abnormal and normal EEG data, including EEG data for Alzheimer's disease, Parkinson's disease, MTBI, and more. In *section IV*, we overview the process of performing machine learning on EEG. Lastly, in *section V*, we summarized the market's current status in the aspects of the six applications we have in *section II*. Our goal is to comprehensively analyze EEG in both the research and the market stage. Finally, we give a conclusion to our paper, which demonstrates that as more EEG applications and markets open up, more EEG devices will be demanded. This demand would inspire more machine learning technologies to be developed. Better machine learning technologies would in turn open up more EEG markets. This iterative process would push forward both the EEG and machine learning technologies.

## 1.1 Background

EEG stands for electroencephalogram, which collects brainwaves from the brain using small metal discs or electrodes attached to patients' scalp [1]. Brain cells communicate using electrical impulses, and electrodes capture these impulses and record them on wavy lines. Brain waves are the rhythmic electrical impulses in the brain when the neurons communicate. Neurons help transport an individual's behavior, emotions, and thoughts within the brain. Brain waves can reveal crucial information about one's general brain function, including but not limited to stress levels, thought patterns, and emotions.

The EEG brain wave can be separated into different bands: delta (0.5-4 Hz), theta (4-8Hz), alpha (8-13 Hz), beta (14-30 Hz), gamma (> 30 Hz), and mu (7.5 - 12.5 Hz). Each band can be and has been used to identify and help researchers with prognostic and diagnostic applications. While no one band is inherently superior to another, some perform better on data analysis for different applications.

German Neurologist Hans Berger first discovered alpha waves, widely regarded as the "father of EEG." Along with beta waves, alpha waves were among the first brain waves documented by Berger, and he displayed curiosity about "alpha blockage," the process where the alpha rhythm is reduced due to visual, auditory, tactile, or cognitive stimulus [2]. In literature, alpha waves indicate the degree of cortical activation, with them being the strongest in the back of the head. Alpha waves gained widespread recognition with the advent of biofeedback theory. Alpha wave biofeedback gained further interest for having some successes in humans as a treatment for depression and seizure suppression [3]. Recently, alpha waves have been used to tackle the science fiction challenge of psychokinesis or the control of the movement of an object using the human brain. In 2009, alpha brain waves were used in a brain-computer interface to control a robotic device [4].

Hans Berger also discovered beta waves in 1924. While the larger amplitude, slower frequency waves were coined the name "alpha waves," the smaller amplitude, faster frequency waves were termed "beta waves." Beta waves are mainly associated with conscious and logical thinking [5,6].

Gamma waves are fast oscillations typically associated with large-scale brain network activity and conscious perception. However, due to its small amplitude and high contamination from muscle artifacts, gamma waves are not widely studied compared to other slow brain waves. Nevertheless, research supports that gamma waves are involved in attention, working memory, and perceptual grouping. Gamma activity has also been observed in many psychiatric disorders such as Alzheimer's disease, epilepsy, and schizophrenia. Among the first reports of gamma wave activity was from a recording from the visual cortex of awake monkeys [7].

W. Grey Walter first described Delta waves in the 1930s, who improved on Hans Berger's research and machine. Delta waves are the slowest and highest amplitude described brainwaves, although recent studies have described slower (<0.1 Hz).

## 1.2 Early History of EEG

EEG is not a novel invention. EEG was invented in the early 1900s. From 1875 to 1890, scientists from the UK found the electrical current in the brain, and researchers from Poland formally defined the brain waves. In 1929, Hans Berger, a German neurologist, developed the first human EEG. Over thirty years after Hans Berger's invention, EEG was mainly inspected visually. From 1961 to 1974, a group of researchers from UCLA started the world of quantitative EEG (qEEG). They used computers to apply Fourier analysis to extract the spectral or frequency features from the EEG signal. From 1974 to the present, EEG began to apply to various fields. For instance, EEG was used by NASA to monitor the brain waves of astronauts on a flight. Engineers developed EEG-controlling robots. Researchers experimented with the process of thought sharing using EEG [8, 9]. Figure 1.2 shows a detailed EEG timeline from 1875 to 2019.

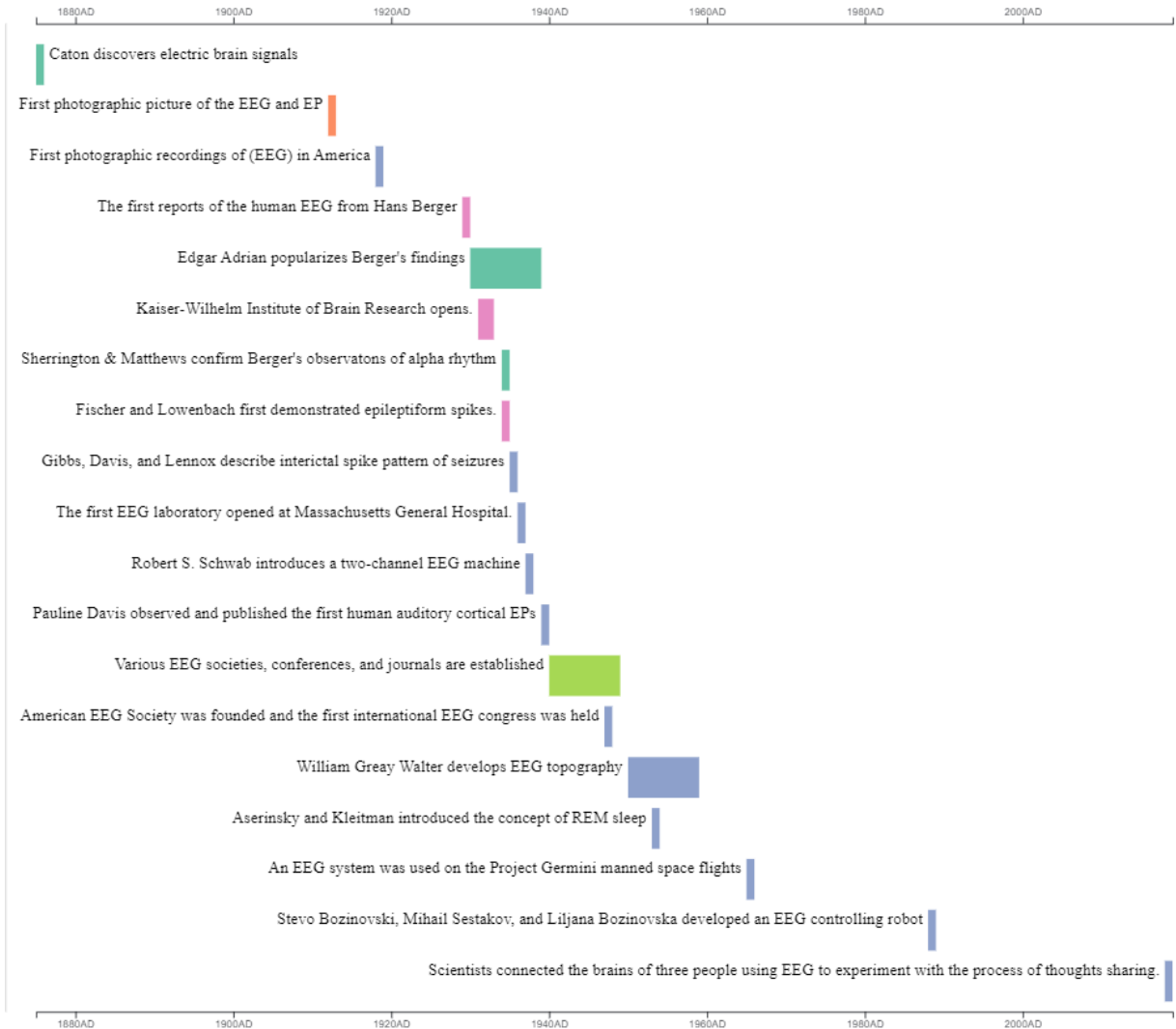

Figure 1.2 The History of EEG

## 1.3 EEG Acquisition

In an EEG machine, there are mainly five processing components: 1) electrodes, 2) amplifier, 3) filter, 4) digitizing, and 5) storage. These five components combine to form a well-performing EEG machine.

1) Electrodes: Electrodes are the first and one of the most critical components. Electrodes capture the electrical impulses from the brain. Therefore, the electrodes can directly affect the quality of the data. Doctors use different types of EEG electrodes in different circumstances. There are two major categories of electrodes: wet and dry.

Dry electrodes are just the common electrodes. Because it is dry, the electrode can create a high interfacial impedance, drastically decreasing the data quality. However, because they are easy to use and do not need too much preparation, Dry electrodes are used in the most common modern EEG system.

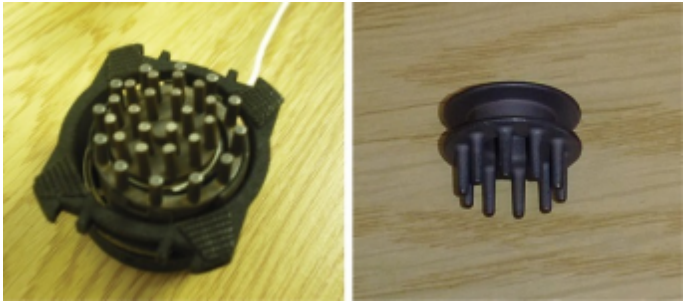

Figure 1.3.1 Dry Electrode [10]

In contrast with the dry electrodes are wet electrodes. Wet EEG is the gold standard of electrodes as it contains gels to reduce impedance; therefore, wet EEG has relatively lower noises and cleaner recordings. However, it also has many limitations. Wet EEG is prone to degregate or dry; therefore, it is restricted to only short-term monitoring. Furthermore, it requires trained technicians and more time to set up. Lastly, the gel can also cause skin irritation.

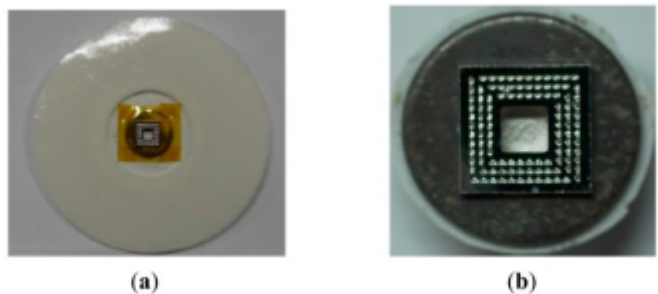

Figure 1.3.2 Wet EEG [12]

As technology continues improving, many special types of electrodes are invented. For instance, a special wet EEG electrode uses solid gel. This electrode improves the drying problem through the hydrophilic hydrogel network [13]. Furthermore, the incorporation of ionic liquids (ILs) into the wet electrode makes the electrodes long-term lasting, although the long-term toxicity of the chemical substance remains a concern [14].

Lastly, the combination of dry and wet electrodes, semi-dry electrodes, has become a popular choice. This is because it minimizes the area exposed to the air, which makes it last longer than wet electrodes. On the other hand, because it is semi-dry, the contact impedance is lower than dry electrodes, which significantly improves the signal-to-noise ratio. Also, because it uses less gel, the semi-dry improved user comforts compared to wet electrodes [15]. In short, the semi-dry electrodes are a combination of dry and wet and can potentially become the golden standard in the future.

**2) Amplifier:** The EEG amplifier amplifies and converts the analog electrical signals from the sensor into digital signals that the computer can process.

**3) Filter:** Digital filtering filters out unwanted signals, mostly noises. The common practice is to apply a low-pass filter and a high-pass filter. The low-pass filter filters out frequencies less than 1Hz, which are not in the range of the human electrical impulses and are considered noises. The high-pass filter filters noise above 40 or 50 Hz [16].

**4) Digitizing:** The EEG digital converter converts the EEG data from analog to digital domain so that the data can be processed into the computer.

**5) Storage:** There needs to be a space to store the EEG data after acquiring it from the human and before the data gets transferred to other places.

## 1.4 Advantages and Disadvantages

There are three main advantages of EEG. First, an EEG device collects data through electrodes. The number of electrodes on EEG devices can be adjusted based on the need, which means the size, price, and complexity of the EEG device is very flexible. The smallest EEG device can be as small as an Airpods, while the larger EEG device can be equivalent to a desktop computer. The price of EEG also can vary from hundreds to thousands of dollars. We have collected data about several EEG devices in *section 5.3 EEG Devices*. The EEG device can contain from 1 electrode to ten or twenty electrodes. Because of its flexibility, the EEG acquisition system can be embedded into many devices, from headphones to VR devices (See our EEG Company Spreadsheet for more information). This allows EEG to apply to many fields, from diagnosis of neurological disorders, long-term monitoring prognosis, and drugs.

The second advantage of EEG is its high temporal resolution, which can measure brain activity in milliseconds [20]. This enables EEG to detect any minor brain disturbances and problems in the early stages. This allows EEG to benefit the medical fields: from the prognosis of diseases to daily health regulation.

Lastly, EEG is non-invasive. Therefore, it does not cause any harm to the human body [22]. This enables EEG for long-term monitoring: depression warning, the prognosis of diseases, and sleep scoring. Also, this property allows EEG to be used on nearly everyone.

However, there is also a critical limitation of EEG. EEG has a relatively low spatial resolution. Because EEG monitors activity in large groups of neurons, it is difficult to pinpoint activity to a

precise location in the brain [21]. EEG can tell "when" an activity occurs, but it cannot tell you "where."

## 1.5 EEG Applications

EEG has many unique properties. For example, EEG has high temporal resolutions, which allows it to detect small changes in the brain. On the other hand, quantitative EEG uses mathematical processing on EEG data, which minimizes the need for electrodes. This makes EEG devices cheap, portable, and easy to use. These unique properties give EEG a wide use of applications in real-life: from disease diagnoses, drug development, neuromarketing, disability assistance and rehabilitation, prognosis, and daily-emotional check, to the future metaverse.

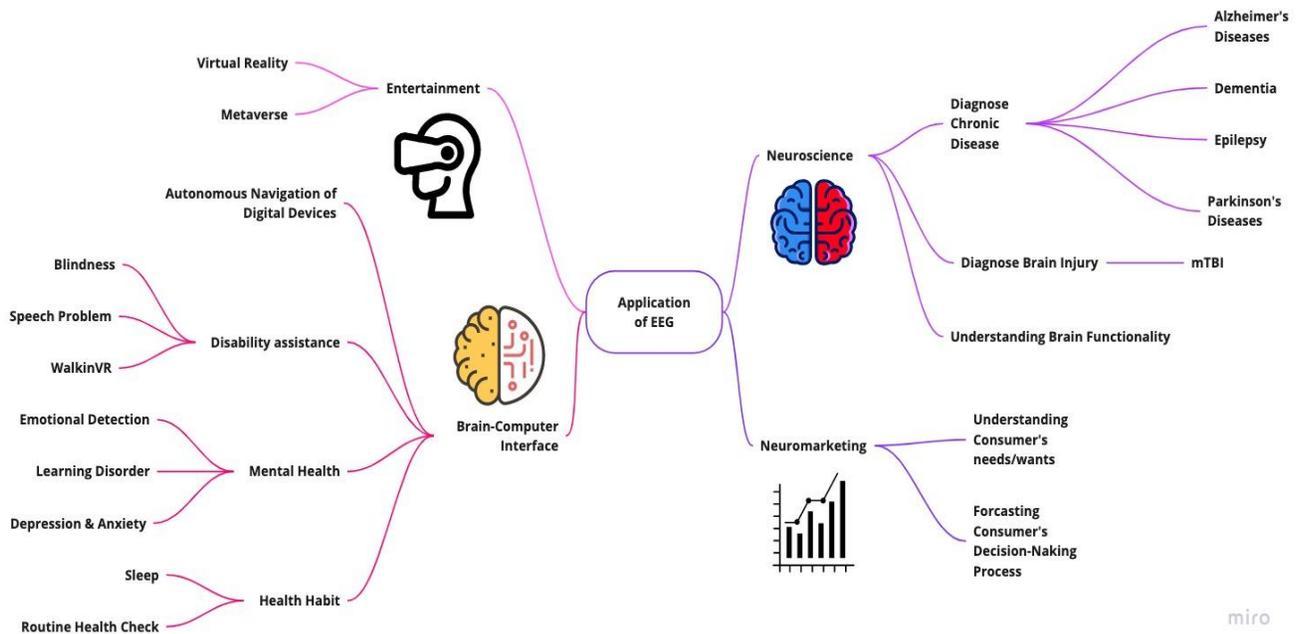

Figure 1.5 Brain Map of the Application of EEG

# II. EEG applications

This section of the paper will cover state-of-the-art EEG applications and up-to-date research studies. This section will summarize the papers in six different categories: 1) Disability Assistance and Rehabilitation, 2) Neuromarketing, 3) Prognosis and Daily Health Check, 4) Metaverse, 5) Drug Development, and 6) Disease diagnoses. All of the papers are after 2017.

## 2.1 Diagnosis/Screen

First, EEG can be used to diagnose neurological diseases. It is already widely used as a tool to diagnose epilepsy. If the EEG of a patient has spikes and sharp waves, which are so-called "epilepsy waves" (figure 2.1.1), this is an indication of epilepsy [23].

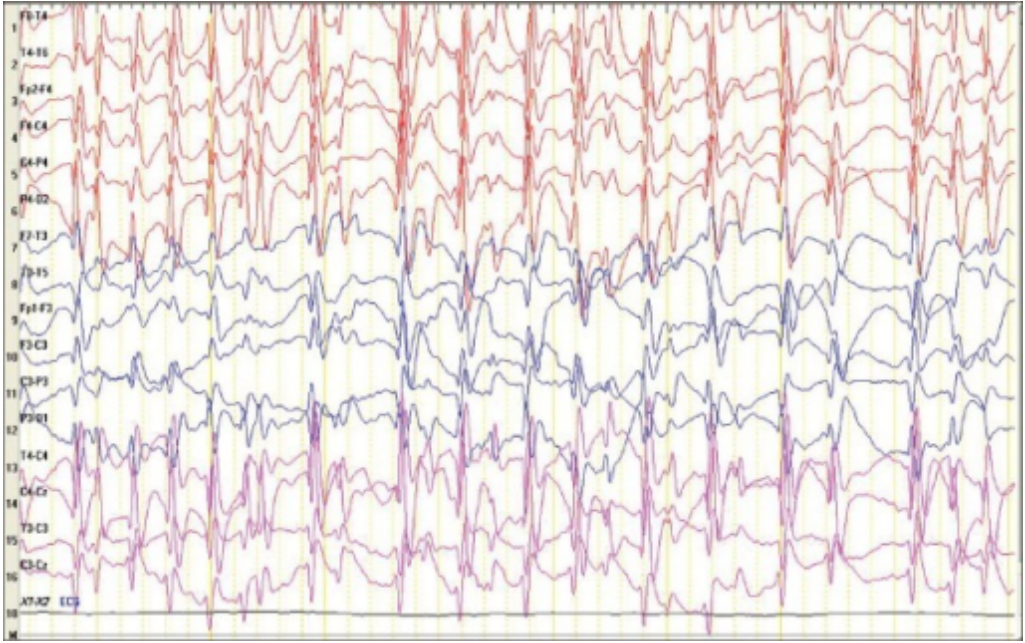

Figure 2.1.1 Epilepsy Spikes [24]

Several studies also indicate EEG's potential for the diagnosis of diseases such as Mild Traumatic Brain Injury (mTBI), Alzheimer's Disease (AD), and Parkinson's disease (PD) [17],[18],[19]. For example, on EEG data, Pirrone performs supervised machine learning, such as Decision Tree, Support Vector Machine, and K-Nearest Neighbor, and reaches 97% when performing binary classifications between healthy controls and AD patients. Lee also performs convolutional-recurrent neural networks on EEG data, resulting in a 99.2% accuracy when classifying PD over healthy controls.

We selected six studies from our other survey paper, "Machine Learning-Based EEG Analysis for Mild Traumatic Brain Injury: A survey," to be included in this subsection. Please note that the six papers we covered in this section are discussed in detail in the other survey paper [25-30]. Instead, we will discuss the trends we found, some limitations of this application, and some suggestions we have. Due to the focus of the other survey paper, all six papers use qEEG data to train machine learning models to diagnose mTBI. All of the papers use some mathematical processing methods, such as discrete Fourier transform, wavelet transform, and others. Dhillon uses mouse EEG [26], while others use human EEG. By comparing the accuracy and the machine learning models, this survey shows that the combination of deep learning and support vector machines proves to be adequate to attain a high accuracy result.

Although all of the accuracies in the table are relatively high, and Lai's model attained 100% [25], most of the studies are binary classification: the models classify between normal and abnormal. Therefore, these models' ability to classify different neurological diseases, such as mTBI and AD, is still unknown. For example, in one of the studies we previously discussed [18], the model attains 97% and 95% when performing AD vs. Healthy or Mild Cognitive Impairment (MCI) vs. Healthy. However, when the researchers perform classification between two disorders, MCI vs. AD, the accuracy drops to 83%. Furthermore, when researchers perform ternary classification, such as MCI vs. AD vs. healthy, the accuracy drops to 75%. This shows that EEG lacks the specificity to differentiate between disorders when performing classification.

In the survey of EEG application in diagnosing diseases, we conclude that EEG can be used to diagnose epilepsy, but its ability to diagnose other neurological diseases remains in doubt. We suggest researchers should focus on increasing EEG's specificity, that is, its ability to differentiate one disorder over another. More studies on multi-classification diagnoses need to be performed to use this technology to help patients and doctors in hospitals. Although the ability of EEG on multi-classified diagnosis is still in the air, EEG has proven its ability on prognosis or screening due to its high temporal resolution, non-invasive, and fast to use. Lai, Dhillon, and Bazarian all proved that EEG screening of diseases is fast and portable, and they have the potential to be used in real-life situations, such as a quick classification of athletes for them to return-to-play or use during the physical examination to detect any brain abnormalities.

## 2.2 Drug Development

Electroencephalogram has long been used as an assistive tool in drug development. More commonly known as pharmaco-electroencephalogram, the process has gained popularity in clinical trials due to its ability to identify biomarkers in the early stages. Moreover, EEG is vital in inclusion and exclusion criteria for safety and efficacy evaluations [31].

We summarized six publications that utilized EEG during the drug development process in detail below:

In 2018
- Koyama experimented on awake rats and tested three types of drugs: pregabalin, EMA401, and minocycline. EEG screened all the mice. The result shows a parallel trend in the reversal of theta power, but not for minocycline. The actual effect of the drug validates the result. Breidenbach suggests that EEG theta power can be used to detect false positive and false negative outcomes of the withdrawal reflex behavior [32].

In 2019

- One study, led by Breidenbach focused on assessing the side effects of drugs. They performed a series of experiments on dogs who have seizures by introducing three different doses. The result shows that EEG can help dose selection to improve safety margin [33].

In 2020
- Lee designed a microfluidic system for continuous EEG monitoring. In their experiment, they recorded EEG from valproic acid (VPA)-treated zebrafish and demonstrated the suppression of seizures, proving their design's effectiveness as a stable screening method [34].

In 2021
- Hua uses qEEG to assess the effectiveness of their new drug ATH-1017's treatment of Alzheimer's disease [35].

In 2022
- Biondi uses transcranial magnetic stimulation (TMS-EEG) to analyze the effects of three anti-epileptic drugs: Lamotrigine, Levetiracetam, and XEN1101. Levetiracetam increased theta, beta, and gamma power. Lamotrigine decreased theta power. XEN1101 increased delta, theta, beta, and gamma power. This result proves that EEG has the potential to indicate the effects of drugs and therefore help doctors to develop the drugs [36].
- Staner investigates how sleep-EEG data can help to develop drugs such as antidepressants and other drugs related to neurological disorders [37].

From our survey, we found that the objective or mainstream of the current EEG drug development is using EEG to assess the effect of their drugs by looking at the increase or decrease of the EEG power bands. While there is substantial research on the topic, there are minimal commercial and clinical uses. In most studies, EEG or qEEG was used as a guiding tool in tandem with other forms of testing.

## 2.3 Disability Assistance and Rehabilitation

Furthermore, integrating EEG into the Brain-Computer Interface (BCI) can help rehabilitation and disability assistance for blindness, speech disorder, motor disability, and other disabilities. In 2018, Dev's team designed an EEG-based brain-controlled wheelchair for quadriplegic patients [39]. Quadriplegic patients cannot move any organs below their necks. In this way, quadriplegic patients can still travel using a wheelchair with their attention level. In the same year, Souza created a virtual environment-based system supported by EEG to help blind wheelchair users [40]. By integrating EEG and Virtual Reality (VR) with 3D sounds into the system, blind wheelchair users can practice in a safe environment until they are used to identifying their location and surroundings with other senses. In the same year, Mroczkowska confirmed using EEG-Neurofeedback (EEG-NFB) to rehabilitate speech disorders in patients after stroke. Their experiment proves that EEG-NFB can be more effective in helping stroke survivors regain verbal fluency than traditional neurophysiological therapy [41].

We summarized ten recently published papers about EEG application on disability assistance and rehabilitation in detail below.

In 2017
- Zhang proposes a 7-layer deep learning model for intent recognition. The result attains a 95.53% accuracy on intent recognition. Furthermore, the model can be built into robots or home automation to help elderly or motor-disabled people control smart devices [38].

In 2018
- An EEG-based Brain Controlled Wheelchair is designed to help the movement of quadriplegic patients. These patients can operate the wheelchair through their attention level and eye-blink [39].
- Souza designed a wheelchair (Figure 1.4.4) adapted with the support of VR and EEG for the training of locomotion and individualized interaction of wheelchair users with visual impairment. The VR provides 3D sounds to stimulate real-life scenarios for blind users. Within the wheelchair, a multilayer computer rehabilitation system was developed that incorporated natural interaction supported by EEG [40].
- Mroczkows separated 58 ischemic stroke patients into experimental and control groups. All patients received neuropsychological therapy, while the experimental group received additional EEG-NFB therapy. After testing the verbal fluency, the result shows that the combination of EEG-NFB therapy with neuropsychological therapy can improve stroke patients' verbal fluency [41].
- Nafea created a brainwave-controlled system that allows the user to switch on and off home applications using blinking and attention levels. This creation was intended to help the disabled and elders and proves the potential of EEG applying for disability assistance [42].
- A group of researchers designed a meal assistance robot to feed the disabled and elders food. First, they used EEG Steady State Visually Evoked Potential to allow the user to identify the food to select. Then, a visual servoing algorithm localized the position of the mouth. Lastly, a mouth open/close status detection system measures the user's willingness to eat at the moment. Fifteen subjects in different experiments are used to validate the design [43].

In 2019
- Casey developed a BCI-controlled robotic arm to help disabled people in their daily life. The team tried two sources of data: EEG and EMG. The results showed that EMG is a very reliable source of signals. However, EEG contains sufficient detection of signals and is not ready to use in clinical trials. This paper points to a limitation on using EEG for disability assistance [44].
- Jacob proposed Artificial Muscle Intelligence with Deep Learning (AMIDL) system to help the movement and communication of stroke patients. The system transforms EEG into body movements by the microcontroller and Transcutaneous Electrical Nerve Stimulation, and the movements are executed by artificial muscle movements. The system also allows communication-based on specific movements that EEG indicates through an online gesture recognition algorithm [45].

In 2021
- Parios studied 23 paralyzed chronic stroke patients and compared their motor cortex activation with the control groups. They found a significant difference between severe and moderate motor deficits in the EEG alpha-beta bands power ratio of the contralesional hemisphere while moving the paralyzed hands. This finding shows the potential of EEG to monitor patient recovery progress [46].

In 2022
- Chinta proposed an imagined world prediction system using CNN to analyze the EEG data. CNN can recognize the words imagined by the patients through visual stimuli. Their team used Morlet Continuous wavelet transform to process the data and attained a 90.3% using CNN model Alexnet. This technology can help people with speech disabilities fully communicate with others [47].

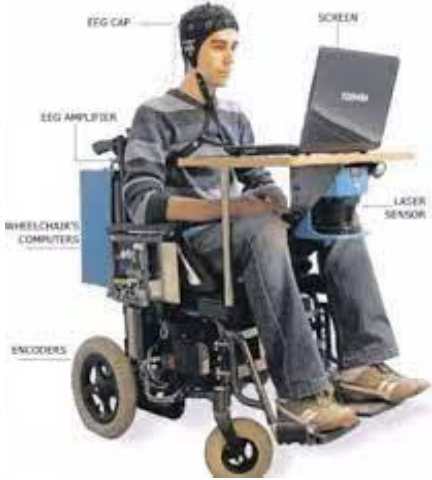
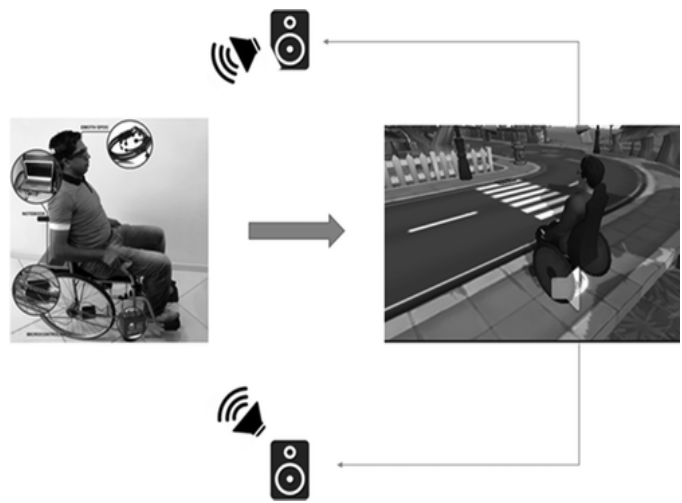

Figure 1.4.3 EEG-controlled wheelchair [48]

Figure 1.4.4 EEG Virtual Environment wheelchair for blind users [40]

There are two main directions of the research: disability of communication or movement. We reviewed the research from 2017 to 2022, and this research shows that the application is progressing in a promising direction.

However, there are still some disagreements. In their study in 2019, Casey showed that EMG is a more reliable method for assistive technology control than EEG. They claimed that EEG showed insufficient signal detection and was inadequate for disability assistance [44]. Lazarou also stated that most current published works do not have sufficient clinical evidence of daily use by disabled people.

From our survey, there is enough evidence to show that the application of disability assistance has sufficient research studies; people created specific EEG-based devices to target a group of disabled people. However, there is still debate on the usefulness of EEG-based disability assistance devices when applied in real life. In the future, researchers should focus on expanding the detected signals and perform their studies with clinical trials in daily settings.

## 2.4 Neuromarketing

EEG also has an application in neuromarketing. It can be used to predict consumers' decision-making process. In detail, we summarized six current published papers about EEG application on neuromarketing below. Most of these papers focus on consumer preferences, which is "like" or "dislike" a specific product, brand, etc.

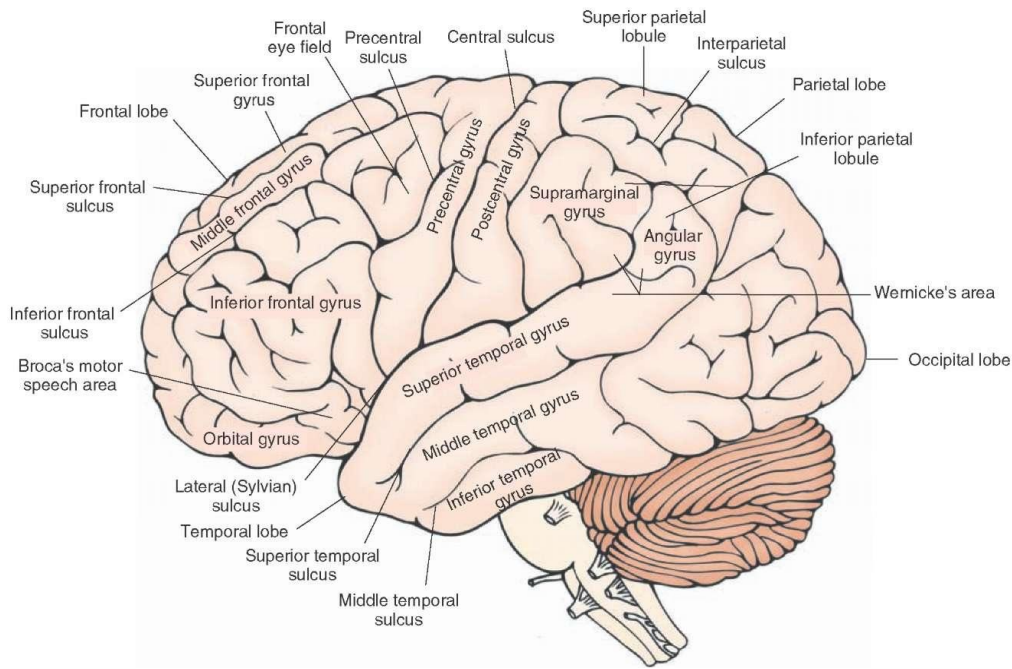

Figure 1.4.2 Frontal and Centro-parietal lobes [49]

In 2017

- Mashrur proposed a MarketBrain prediction system to predict consumer preferences. First, they recorded the EEG while showing the volunteers five product endorsements. Then, they used Wavelet Packet Transform and SVM, achieving a 96.01% accuracy [50].

In 2019
- In their 2019 experiment, Molnar-Nik's group used EEG data to analyze consumer preferences in the advertisements for different phone brands. From EEG power, Golnar-Nik located the frontal and Centro-parietal locations (Figure 1.4.2) to be the most critical in predicting the "like" and "dislike" decisions of consumers. In the end, they achieved an 87% in predicting consumer decision-making incidence [55].

- Ciorciari explores the relationship between emotional intelligence, commercial advertising messages, and consumer behaviors. The study recorded EEG from 45 participants while the participants were watching a series of ads. The study claims that different types of advertising messages can attract people with different levels of emotional intelligence [51].

In 2020
- Sun conducts a study on the consumer decision-making of online shopping. The EEG is recorded from volunteers watching a series of products and prices. After analyzing it, Sun found that rating is more crucial in influencing consumers' online purchasing decisions than the price [52].
- Aldayel performs an EEG-based preference recognition study and uses different ML models to predict consumers' behavior on specific products based on EEG data. The result shows that the DL model attained the highest accuracy, 94%, compared to other ML models [53.]

In 2022
- Phutela researches to understand consumers' positive and negative reactions to ads and products by analyzing EEG signals from the volunteers while they are watching different ads. ML models such as Naive Bayes (NB), SVM, K-Nearest Neighbor (KNN), and deep learning (DL) are used to predict consumers' reactions. As a result, SVM and DL have the highest accuracy at 56% [54].

Three of the six papers [50, 53, 54] utilized ML models to create intelligent consumer behavior prediction systems. Aldayel and Mashrur achieve high accuracies: 94% and 96.01%, respectively. In addition, both papers prove that SVM is the most effective tool in classifying consumers' like or dislike; SVM achieves 90% or higher in both papers. However, Phutela uses the Naive Bayes Classifier but only attains 61% in subject-dependent analysis.

The majority of the papers prove that EEG has the potential to be used in neuromarketing in their experiments. However, there is a lack of evidence of the actual effectiveness of EEG when used in real-life situations: whether the companies' sales can increase by using these EEG consumers' behavior analysis. Therefore, we suggest that researchers do more studies to validate the effectiveness and consistency of EEG in neuromarketing and its usefulness in the market.

## 2.5 Daily Health

EEG has an excellent capability to use as a tool of prognosis due to its high temporal resolution. This makes EEG enable to detect small disturbances in the brain. In recent years, EEG has cemented its spot as the most widely used prognostic tool in clinical examinations and is accessible both in hospitals and at home; it is used in prognostication and ruling out subclinical seizures [76].

We summarized a total of 5 papers that utilized EEG as a tool to interpret and analyze data in order to continue health checks. We reviewed each paper in detail below.

- In 2018
  - Cai focused their study on the automatic classifying of depression. The EEG data is acquired from 213 subjects. A finite Impulse Response filter combining the Kalman derivation formula is used to denoise the data. Discrete Wavelet Transformation and Adaptive Predictor Filter extracts 270 linear and nonlinear features. These features are then sent into the minimal-redundancy-maximal-relevance (MRMR) feature selection technique to reduce the dimensionality of the features. Lastly, SVM, KNN, Classification Tree, and Artificial Neural Network are used to classify depression. KNN has the highest accuracy of 79.27% [58].
- In 2019
  - Mousavi proposed an automatic sleep stage annotation method, SleepEEGNet, composed of CNNs. Time-variant features, frequency information, and long short-term context between sleep epochs and scores are extracted as features. They also applied a loss function to avoid class imbalance problems—their method results in 84.26% [60].
- In 2020
  - Melentev used EEG and ML methods to classify the professional level and tiredness of the eSports players. After filtering and removing the artifacts, they used statistical differences before and after the game as features for two binary classification problems: player professionalism and health rate. They result in a 95% F1-score to detect professional players and a 90% F1-score to classify player health [61].
- In 2021
  - Satapathy proposed an ensemble learning stacking model (ELSM) using EEG data to classify sleep stages. They extracted 28 features and used the Relief-based Weight Feature Selection Algorithm to screen features and Pearson correlation coefficients to eliminate redundant features. Their proposed ELSM achieved 99.34%, 90.8%, and 98.5% in three different datasets testing [60].
- In 2022
  - Liu performs a study on the classification of major depressive disorder. The EEG data are acquired from 29 healthy subjects and 24 patients with severe depression. Using the proposed end-to-end deep learning framework results in an accuracy of 90.98% [62].

Cai and Melentev focused their studies on depression; Mousavi and Satapathy focused on sleeping; Suhaimi focused on tiredness. Nearly all papers we collected utilized some form of machine learning to process the EEG data. Among these papers, Satapathy's ELSM classifying sleep stages and Suhaimi's method classifying professional level and tiredness of the eSports players seem promising.

## 2.6 Metaverse

Metaverse is a novel term that has come to people's vision in recent years. As metaverse became popular, related technologies, such as AR and VR, also got much attention. As a result, there has been a recent increase in applying EEG devices and research towards VR and AR technologies, especially by large tech companies such as Google, Apple, and Meta.

We collected seven papers that implemented EEG and focused on developing AR or VR-related products. We reviewed each paper in detail below.

In 2019
- Vortmann focused on solving one problem in AR applications: the interference of virtual objects with the user's current attentional focus. Thus, an EEG-based classification is proposed using the Linear Discriminant Analysis classifier. As a result, they achieved 85.37% accuracy in classifying the virtual objects in AR and the current attention of the user [64].

In 2021
- Amoolya created an emotional detection system based on EEG signals. The most notable feature of this work is the creation of a readily accessible and easy-to-use App with AR. This system has the potential to embed into the future metaverse and allows each user to express their emotions on their virtual character [65].

In 2022
- Liu designed a ResNet-based multi-feature fusion method for motor imagery task classification. To address the limitation of EEG, such as insufficient information or low signal-to-noise ratio, Liu used three time-frequency transforms, including morlet wavelet transform, multi-window time-frequency transform, and stockwell transform to extract features. The features are then sent into ResNets for further feature extraction and classification. The model results in 97.86% accuracy [66].

- In 2022, researchers Guo Honyu and Gao Wurong constructed an experiential situational English-teaching scenario and convolutional neural networks (CNNs)-recurrent neural networks (RNNs) fusion models to recognize students' emotion EEG in experiential English teaching during the feature space of time domain, frequency domain, and spatial domain [67].

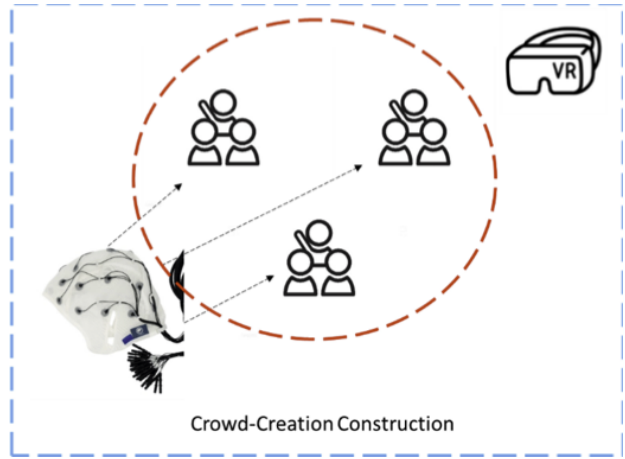

Figure 1.4.5 Metaverse-Powered Experiential Situational English-teaching [67]

- Lee developed a brain-to-speech (BTS) system using EEG for real-world smart communication using brain signals. They proposed pseudo-online analysis and achieved 47% when the chance level was 7.7% and 76% when it was 50% [71].

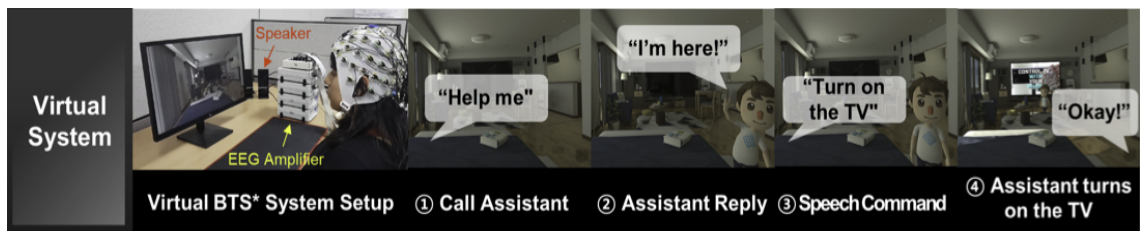

Figure 1.4.6 Imagined-Speech-Based-Smart-Communication System [71]

- Seok Hee Oh aimed to verify the effectiveness of VR cognitive training by measuring the sense of presence and EEG in children with ADHD. However, after running various experiments, they concluded that EEG could only be used as an auxiliary means as no significant EEG changes were observed after VR cognitive training [72].

Our survey found many exciting innovations, from an emotional detection system, motor imagery task classification, and experiential English Teaching System to a Brain-to-speech system. As these technologies mature and become more advanced, more functions will be added to the virtual world, and one day, humans will be able to achieve the true metaverse. However, for now, more research and AR and VR studies are required.

## III. EEG Data

During the process of our literature review, most papers did not publish the EEG datasets that were used in their research. Moreover, EEG data for certain diseases are especially difficult to access or find given the Health Insurance Portability and Accountability Act, or HIPPA, which prevents sensitive patient data from being disclosed without the patient's knowledge or consent. Nevertheless, this data is vital in machine learning and artificial intelligence research. Both normal and abnormal EEG signals must be present to prevent biases in diagnostic ML and AI software. In this section, we have collected twelve various EEG datasets: including seizures, Parkinson's disease, Alzheimer's disease, depression, sleep disorder, mTBI, alcohol, and three normal public EEG data repositories. We hope our data collection can inspire and help other researchers who are interested in research but do not have EEG data.

While these datasets are available to the public, there are still many diseases and scenarios where EEG data is scarce. Diseases such as Epilepsy, Motor Imagery, or Seizures are among the most flushed out and well-documented illnesses by EEG. However, illnesses such as mild traumatic brain injury severely lack publicly available EEG datasets. The lack of publicly available EEG datasets is ironic, given that hundreds of thousands of EEGs are recorded daily across the country and globe. Furthermore, A relatively small amount of this data is publicly available to the research community in the form of meaningful and usable machine learning data. Massive amounts of public EEG data would allow researchers to use advanced machine learning algorithms to uncover new diagnostic methods and confirm clinical practices. Additionally, data collected in clinical settings rather than tightly controlled research environments are preferred. This is because "clinical-grade" data is intrinsically more variable with respect to parameters such as environment, electrode location, mood, noise, etc. Recording this variability is vital for developing long-lasting, high-performance technology with viable real-world applications [77].

| Name | Disease | Volunteers | Data Size | Data File Type |
| --- | --- | --- | --- | --- |
| JasonChiehLee | concussion | 20 patients | | matlab |
| Alzheimer's Classification | Alzheimer's Disease & Mild Cognitive Impairment | 80 people (normal vs. AD vs. MCI) | | csv |
| Diagnosis of Alzheimer's disease with Electroencephalography in a differential framework | Alzheimer's DIsease | 169 subjects | 30-channelEEG at 256Hz at rest with close eyes in 20 minutes | ... |
| TUH Abnormal EEG Corpus | Seizure and Alzheimer's Disease | Unknown | ... | ... |
| Predict | 7 Parkinson's Disease, 3 mTBI, 1 OCD, 2 Depression, and 1 Schizophrenia | total 496 patients | unknown | varies |
| National Sleep Research Source | Sleep Apnea & a variety of other sleep EEG recordings | Multiple datasets | Varies by dataset | Varies |
| TDBrain Clinical Lifespan Database | ADHD, OCD, Depression, and Subjective memory complaints | 1247 | 26 channel, 2 minutes eye open, 2 minutes eye closed | |
| CHB-MIT Scalp EEG Database | Intracable seizures | 22(5 males, ages 3–22; and 17 females, ages 1.5–19) | 23 signal(24 or 26 for rare cases) | ZIP file |
| Continous EEG | alcoholic patients | 122 | 64 channel at 256 Hz for one second | unknown |
| Narayanan Lab | Multiple but more Parkinson's | varies by dataset | varies by dataset | unknown |
| Open Datasets in Electrophysiology | Motor, imagery, auditory, and many more | varies by dataset | varies by dataset | unknown |
| EEG Datasets | Motor, imagery, auditory, and many more | varies by dataset | varies by dataset | unknown |

# IV. Machine learning using EEG

Due to the complexity of the EEG data, most of the EEG analysis we covered in section II EEG Applications used machine learning to achieve their goal. Machine learning includes data acquisition, processing, feature extraction, and machine learning. We already covered the entire process of EEG data acquisition in *1.3 EEG Acquisition*. The EEG system has five main parts: electrodes, amplifier, filter, digitizing, and storage. After the EEG data is collected, it must go through data processing methods to remove any noises and artifacts. This can be done by visual inspection or statistical methods to remove those outliers. The third part is feature extraction. Feature extraction selects those critical features to later feed into the machine learning models. Standard EEG feature extraction methods include Fourier transform, which can be broken down into either the fast Fourier transform or the discrete Fourier transform. Wavelet analysis is also a prevalent feature extraction method. Also important to note that most deep learning methods, such as convolutional neural networks (CNN) or recurrent neural networks (RNN), can automatically extract features. The last part is machine learning methods. The most common machine learning methods for EEG are summarized below in Figure 4.1. The figure includes 27 machine learning methods, broken down into classification, clustering, deep learning, and dimensionality reduction. In our survey Paper Machine Learning Based EEG Analysis for Mild Traumatic Brain Injury: A survey, we covered a comprehensive survey with state-of-the-arts on EEG machine learning mTBI diagnosis. In that paper, we summarized all four processes described above in detail. The mTBI survey concludes that Support Vector Machine (SVM) is the best machine learning method to diagnose mTBI as it has higher accuracy than other ML methods. Deep learning, such as LSTM and CNN, also shows promising accuracy. Lastly, the mTBI survey pointed out that a combination of SVM with a deep learning method is the best architecture for diagnosing mTBI though the results are usually unexplainable as the features are automatically extracted through layers.

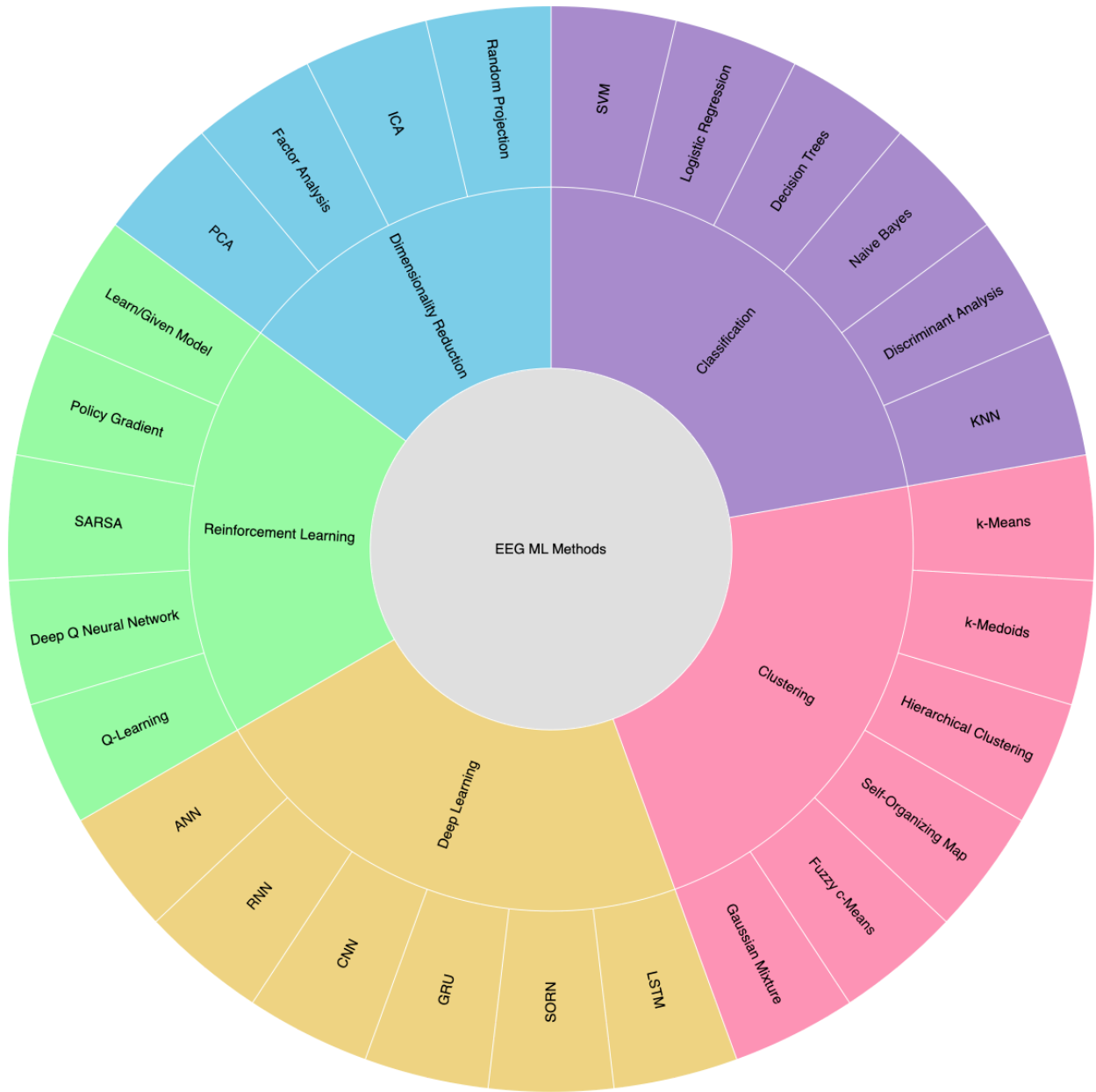

Figure 4.1 EEG Machine Learning Methods

| ANN - Artificial Neural Network | LSTM - Long Short-Term Memory | SARSA - State-Action-Reward-State-Action |
| --- | --- | --- |
| RNN - Recurrent Neural Network | KNN - k Nearest Neighbor | SORN - Side-Output Residual Network |
| CNN - Convolutional Neural Network | SVM - Support Vector Machine | PCA - Principal Component Analysis |
| GRU - Gated Recurrent Unit | ICA - Independent Component | |

# V. Market Analysis

## 5.1 Applications in Market

In our search, we collected 131 companies about EEG. We then separated each company into one or two of six categories: Disease/Screening, Neuromarketing, Drug Development, Disability/Aged Rehabilitation and Assistance, Daily Health, and Metaverse. Each category is a part of the larger EEG ecosystem, and while they do not encompass the entirety of the current EEG research and market space, they offer an insight into the current state of the EEG market space.

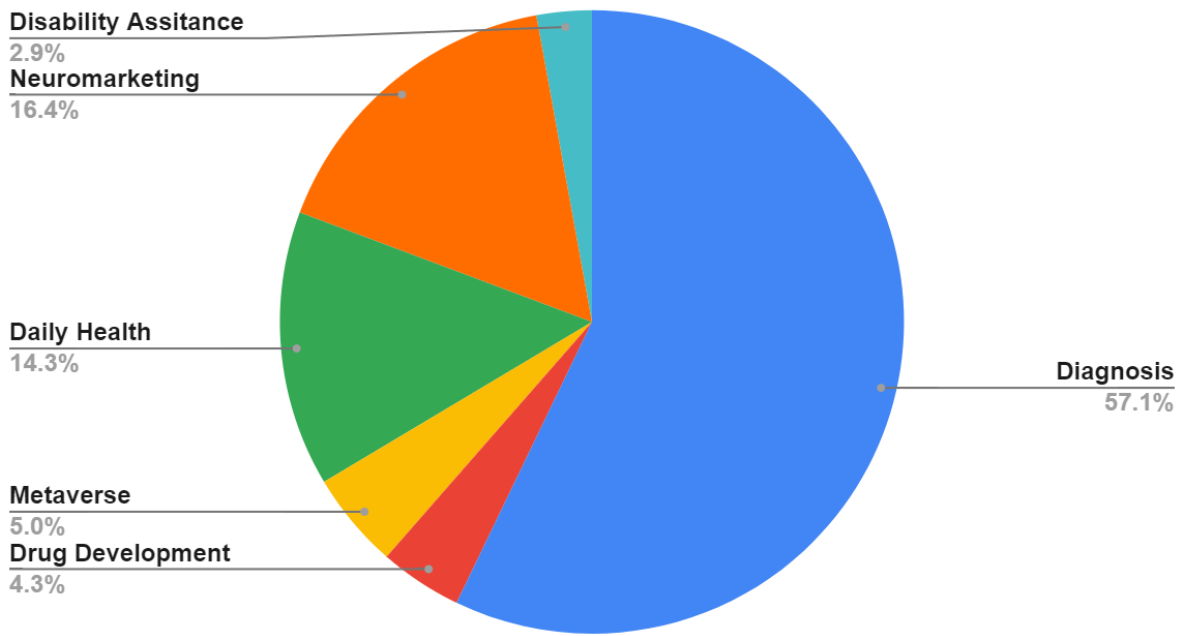

Figure 5.1.1 EEG Company market breakdown

Companies mainly focused on diagnostic services, and devices were categorized into the diagnosis and prognosis categories. Moreover, companies that solely create devices such as caps or amplifiers were also categorized in the diagnosis and prognosis category. Of the 131 companies we found, 80 companies had some degree of involvement in the diagnosis and prognosis. This category was separated into two further subcategories: Devices and Services. Companies that focused on services varied in the solutions they offered. For example, SleepMed

Inc. is a company that focuses on high-quality streamed EEG recording for sleep quality. SleepMed Inc. is just a few companies offering 24/7 ambulatory EEG services. Another common type of service company was companies focused on analyzing EEG data. These companies often have proprietary software that analyzes, cleans, and/or filters raw EEG recordings and 24-hour assistance to address client needs. For example, Neuronostics utilizes its BioEp software to detect epilepsy. They are just one of many companies we collected which employ diagnostic or analytic software to process and make decisions on EEG data. Device companies were also quite prominent in our market search. Companies like BrainScope are well-established and have many FDA-approved EEG devices and software. EEG devices show a wide degree of variation from caps and amplifiers used in research to headbands focused explicitly on specific diseases or illnesses. For instance, Ceribell provides an FDA-approved EEG headband, the Rapid Response EEG, which focuses on quickly diagnosing Epilepsy and Seizures. In Figure 5.1.2, it can be seen that diagnostic services are the more predominant type of EEG diagnosis company. It's also worthy to point out that most of these EEG diagnostic companies are focused on epilepsy and seizures. While not as prominent as EEG diagnosis service companies, EEG diagnosis device companies are still a mature sector of the EEG ecosystem.

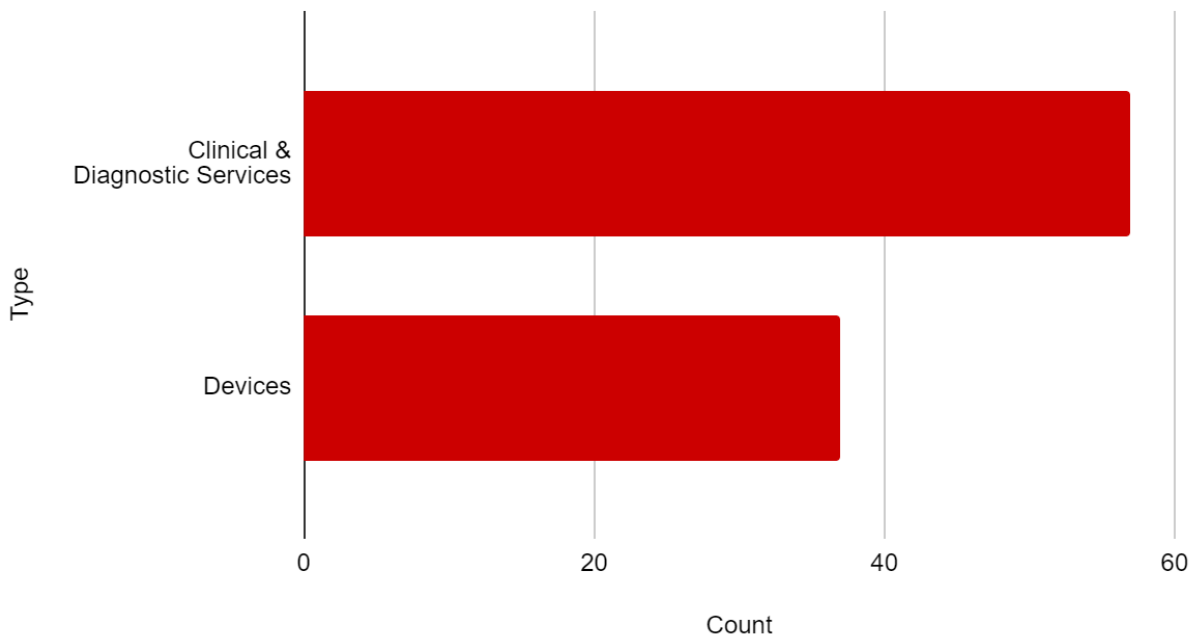

Figure 5.1.2 Disease and Prognosis Company Graph

Neuromarketing is the category with the second highest number of companies. This category mainly focuses on using EEG recordings to aid in the user interface (UI), user experience (UX), advertisement, and more. EEG recordings are used to measure the user's engagement, emotion, and focus, which are then used to help make websites, advertisements, or apps more user-friendly. The scope of neuromarketing is vast and encompasses a broad spectrum of companies, from news websites to e-commerce. Neuromarketing is among the most promising fields of EEG. For example, Brainengineers utilizes EEG to record engagement and user emotion and identify any pain points in UX. Using EEG in marketing has allowed companies, whether Fortune 500 or small businesses, to improve user engagement and experience. While a relatively premature sector of the EEG ecosystem, neuromarketing shows immense potential and serves as the bridge between neuroscience and business.

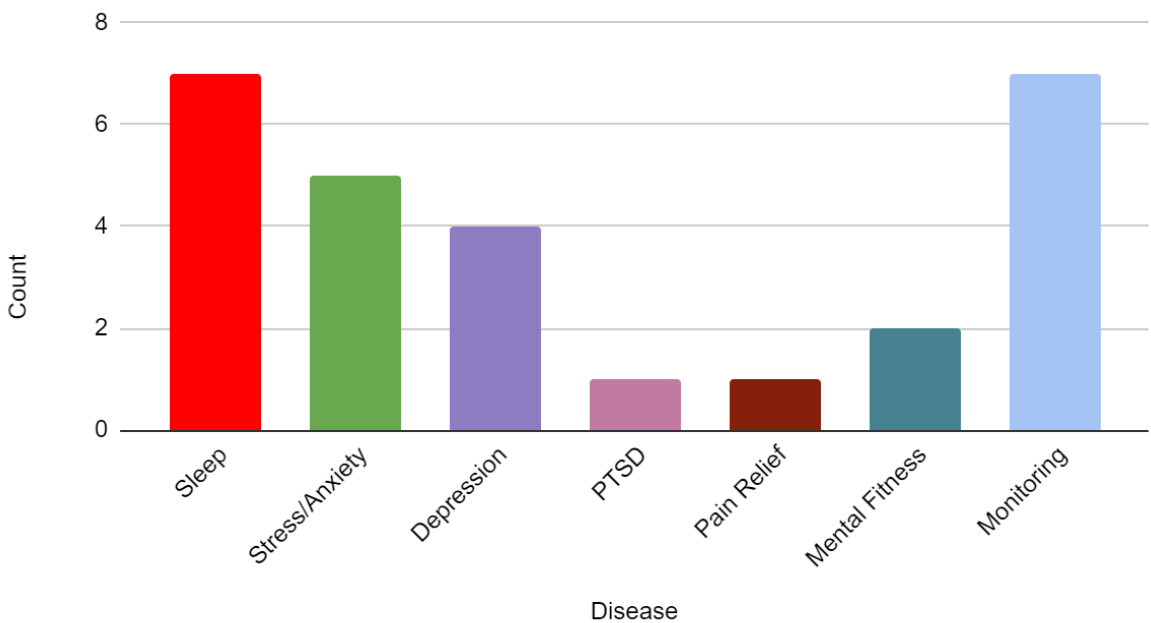

Figure 5.1.3 Daily Health Disease Count Graph

Next, daily health was the third most common company in our market search. Daily health companies were categorized as such if they focused on depression, sleep quality, or other emotional disorders. In Figure 5.1.3, all illnesses that daily health companies addressed were counted and documented. From the figure, daily health companies which focused on monitoring and sleep were the most numerous. These companies focused predominantly on observing and controlling sleep quality and life. Other companies also focus on stress, anxiety, depression, and a wide variety of other mental illnesses. For example, Enophone is a company that offers EEG headphones; eno has embedded EEG sensors to track and monitor mental health and fitness. The EEG recordings allow distractions and other obstacles to be identified and treated while mental performance is optimized. While the daily health category is not as numerous as diagnosis and prognosis or neuromarketing, the field is well-developed, and a broad range of companies address various illnesses in daily life.

Metaverse companies were categorized as such if they focused on virtual reality and augmented reality. In our search, there were only seven companies that met this criterion. Contrary to popular belief, metaverse companies are not all focused on the gaming or entertainment industries; most companies we collected use EEG and VR/AR tech for research. However, some companies still aim to combine EEG and VR to create games. One such company is Neurable, which in 2017 unveiled the first EEG-controlled VR game. Again, however, most companies use EEG and VR as research and development tools.

Drug development companies are companies that utilize EEG as a way to develop and test pharmaceutical drugs. In our search, we only found six drug development companies. These companies mainly focus on utilizing EEG recordings to test and validate drug side effects and their impact on the user's psyche. For example, Beacon Signals is a company that uses machine learning algorithms to analyze EEG and find neuro biomarkers during the drug development process. Most drug development companies are directly involved in the clinical trial process, and the addition of EEG attempts to speed up the usually decades-long process.

Finally, disability rehabilitation and assistance companies produce devices or software to help disabled individuals. This category had the smallest number of companies in our category, which may be attributed to its relatively short development. Most devices are still in the research and development stage, while most companies we found mainly focus on vision or hearing impairments, stroke patients, or other mental disabilities. Physical disabilities such as amputation or cerebral palsy are not as well addressed as other disabilities. While this field is relatively premature, we do find a lot of research papers related to this application in *2.3 Disability Assistance and Rehabilitation* section. Our hypothesis is that the designs in *2.3 Disability Assistance and Rehabilitation* might not be very useful or cost-effective in the business market. Many of those devices might cost more and have higher embedded risks compared to human-care. In addition, this contrast between the research stage and the market stage shows that humans still do not trust robots or automation on life-mattered things.

## 5.2 EEG Device Market

In this survey, we have summarized the 22 most common headsets used to collect EEG data. In our table, the majority of the headsets use dry EEG electrodes, and only two of the headsets use wet EEG electrodes. The Table 5.2 acts as evidence and proves our previous point that dry EEG electrodes are the most common ones in the current market. This abundance is primarily due to its easy-to-use trait.

| Name | dry(1)/wet(0) | weight (grams) | Price($) | Channels | SR(Hz) | SD(bit) |
|---|---|---|---|---|---|---|
| Diadem | 1 | 190 g | ... | 12 | 256 | 24 |
| Air | 1 | 130 g | ... | 8 | 256 | 24 |
| Hero | 1 | 250 g | ... | 9 | 256 | 24 |
| CGX Quick-29R V2 | 1 | 526 g | 20,000.00 | 21 | 500 | 24 |
| CGX-Cleared-Quick-20m | 1 | 596 g | 20,000.00 | | 500 | 24 |
| CGX-New Quick-32r | 1 | 646 g | 20,000.00 | 30 | 500 | 24 |
| CGX Mobile 72 | 0 | 1000 g | 20,000.00 | 72 | 500 | 24 |
| CGX Mobile 128 | 0 | 1000 g | 20,000.00 | 128 | 500 | 24 |
| Wearable Sensing DSI-24 | 1 | 600 g | 24,800.00 | 21 | 300/600 | 8 |
| Wearable Sensing DSI-7 | 1 | 350 g | ... | 7 | 600 | 4 |
| Wearable Sensing DSI-7 Flex | 1 | 200 g | ... | 6 | 600 | 4 |
| Wearable Sensing VR 300 | 1 | 400 g | ... | 7 | 300 | 4 |
| Cognionics Quick 30 | 1 | 250 g | $22,000.00 | 30 | 540 | 24 |
| BrainScope One EEG-based Device | ... | ~600 g | 24,975.00 | 20 | 1000 | 45 |
| DSI-VR300 | 1 | ~400 g | | 7 | 600 | 4 |
| Muse 2 | 1 | 51 g | 200.00 | 4 | 256 | 12 |
| OpenBCI v1 | 1 | ... | 200.00 | 4 | 256 | 24 |
| OpenBCI v2 | 1 | ... | 500.00 | 8 | 256 | 24 |
| OpenBCI v3 | 1 | ... | 945.00 | 16 | 256 | 24 |
| Emotiv Insight | 1 | ... | 300.00 | 5 | 128 | 14 |
| Emotiv Epoc | 1 | 170 g | 850.00 | 14 | 128/256 | 14/16 |
| Neurosky Mindwave | 1 | 90 g | 100.00 | 1 | 512 | 12 |

Table 5.2 EEG devices

## 5.3 Market Size

As the EEG market continues to mature and develop, more and more companies are founded, tackling some of EEG's most significant problems and further refining the field. New EEG devices that are more portable, cheaper, and advanced are appearing on the market. In 2010, the most affordable EEG headset cost $600, whereas they are now as low as $99 [56]. Studying the reliability of these cheaper EEG devices for emotional or other neurological studies would be interesting, as these devices could be more economical and compatible with VR settings. The VR market is witnessing an 8.7% annual increase in market size [57].

In the coming years, the necessity of EEG testing devices will become evident due to lowering device costs and an increasing prevalence of neurological disorders among patients. According to the American Hospital Association (AHA), there are 6,146 hospitals in the United States, with the vast majority either in need or interested in acquiring EEG technology such as monitoring and testing devices. Currently, 88% of universities, 55% of teaching, and 14% of general hospitals use ICU continuous EEG (cEEG) globally [63][68]. EEG usage is increasing globally.

For example, in 2016, ICU cEEG was used in 37% of all hospitals in the Netherlands, a 16% increase compared to the 21% in 2008. [68].

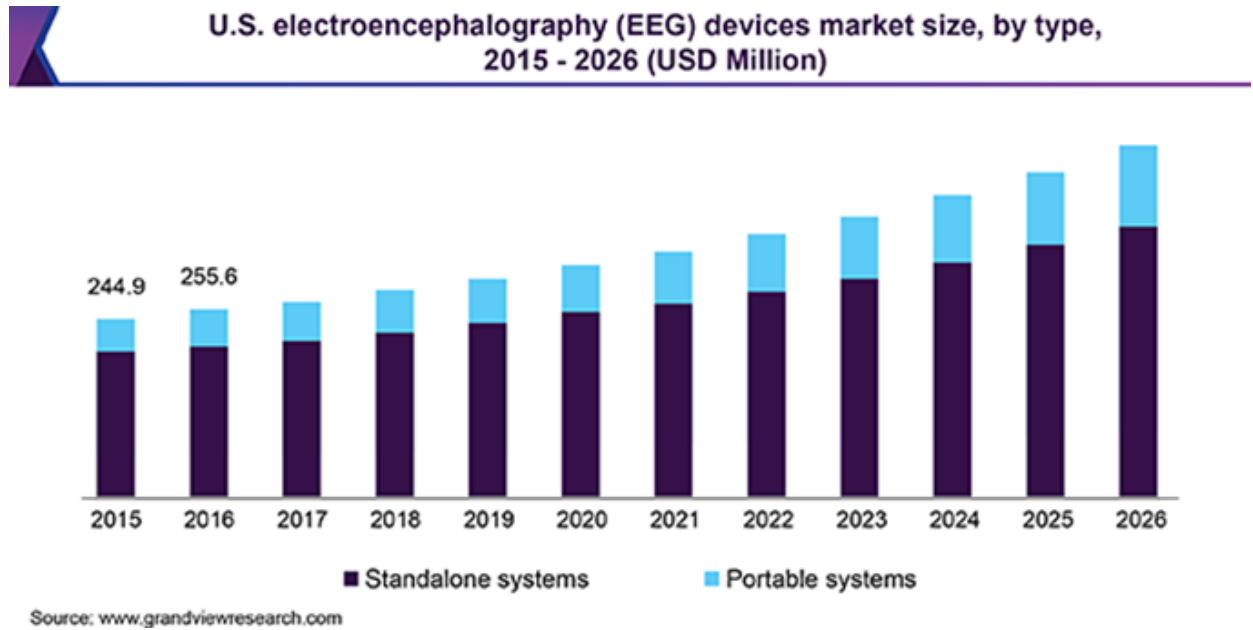

Figure 5.3.1 U.S. EEG devices market size, 2015-2028 [69]

EEG products are expected to have a high penetration rate as the technology is necessary for hospitals and patients. This is due to the overall niche market space EEG technology occupies. Out of all EEG-based companies, there are roughly 14 companies that provide most of the EEG monitoring, testing, and software to hospitals and clinics [75]. While the market is niche and relatively premature, there is a 9% increase in job demand for radiologic and MRI technologists who can operate EEG devices [57]. According to a new study by Report Ocean, the global EEG devices market size is expected to reach USD $2.03 billion by 2028. There are a variety of factors that contribute to this increase. However, a growing incidence of neurovascular disorders and a significantly rising elderly population contribute to market growth [70]. The number of future EEG users can be estimated by looking at the total number of patients with neurological diseases that can be treated with EEG. For example, the total number of people with dementia worldwide was reported at 35.6 million and is expected to double nearly every 20 years to 65.7 million in 2030 and roughly 115.4 million in 2050 [70].

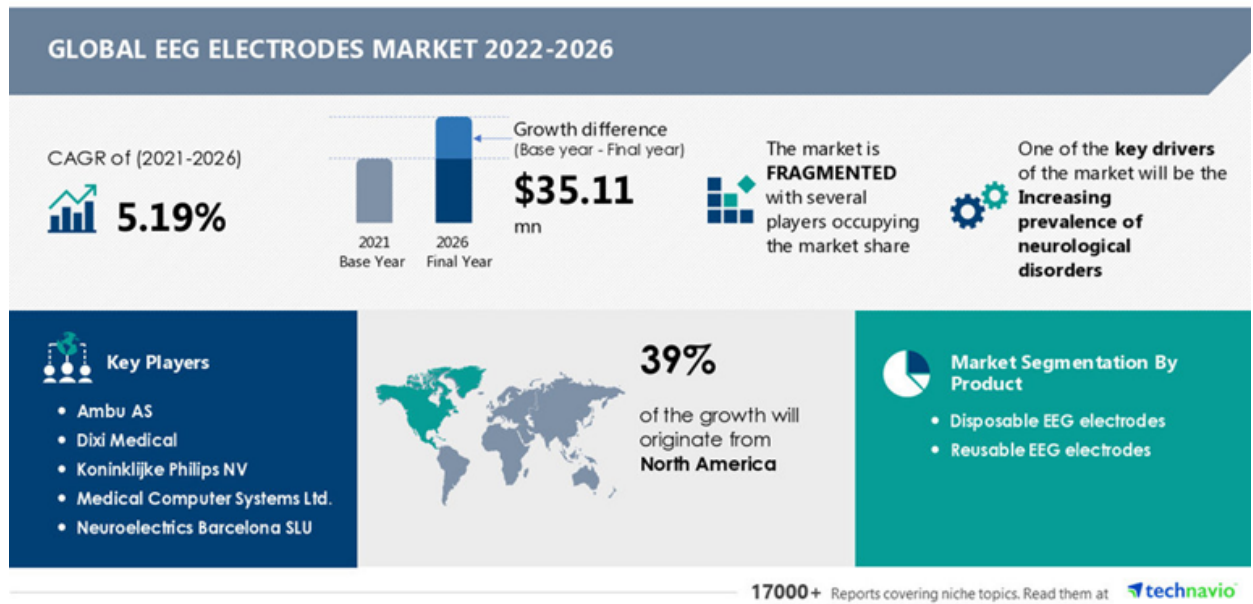

Figure 5.3.2 Attractive Opportunities in EEG Electrodes Market by Product and Geography - Forecast and Analysis 2022-2026 [73]

## VI Conclusion

This paper investigated various applications of EEG-related ecosystems, including EEG research, devices, companies, and markets. We summarized the conditions of the six applications of EEG (diagnosis/screening, neuromarketing, drug development, disability rehabilitation/aged assistance, daily health, and metaverse). The application of drug development, metaverse, and neuromarketing is promising and still has much potential to grow. Due to EEG's inherent trait, we strongly believe in using EEG for disease screening and daily health. EEG's application to diagnosing the disease is still in research. Although EEG-based machine learning performs well in binary classification, EEG lacks the specificity to differentiate between abnormal EEG data and performs poorly on multi-class classification. We found a very different situation in the research field and the market for disability rehabilitation and aged assistance. In the research field, there are various papers with different innovations, while in the market, there are very few companies. This result implies people haven't totally accepted the automation AI. In addition, the cost of the devices might also be a reason. Nevertheless, these six applications have been developed and are growing very quickly, especially in diagnosis/screening and daily health. We expect that there will be more EEG applications and market open up.

Our results show that the EEG market has great potential to grow very big because machine learning technologies advance the EEG market forward, which in turn helps everyone in their daily life. Then the demand for EEG services requires more devices and applications to be developed. This interactive and mutually push-forward force compelled the EEG ecosystem to expand in response.

Further, EEG is being pushed by new applications in brain computer interface, as shown in the new emergent market of metaverse. In addition to those new applications, EEG companies like [Dasion](#) build fast algorithms to provide explainable results and enable multi-class classification with high accuracy.